\documentclass{nextgame2026} 

\usepackage{microtype}
\usepackage{hyperref}
\usepackage{url}
\usepackage{booktabs}
\usepackage{multirow}
\usepackage{enumitem}
\usepackage{amsmath}
\usepackage{listings}
\usepackage{xcolor}
\usepackage{tikz}
\usetikzlibrary{arrows.meta,positioning,fit,backgrounds}
\usepackage{graphicx}

\usepackage{lineno}

\usepackage{amsmath}
\usepackage{amssymb}

\usepackage{booktabs}

\usepackage{float}

\usepackage[T1]{fontenc}

\definecolor{darkblue}{rgb}{0, 0, 0.5}
\hypersetup{colorlinks=true, citecolor=darkblue, linkcolor=darkblue, urlcolor=darkblue}

\lstdefinestyle{pysmall}{
  language=Python,
  basicstyle=\ttfamily\scriptsize,
  keywordstyle=\bfseries\color{blue!70!black},
  commentstyle=\itshape\color{green!50!black},
  stringstyle=\color{red!60!black},
  showstringspaces=false,
  breaklines=true,
  frame=single,
  framerule=0.4pt,
  xleftmargin=2pt,
  xrightmargin=2pt,
  aboveskip=4pt,
  belowskip=4pt,
  numbers=none,
}

\newcommand{\Mllm}{\mathcal{M}}
\newcommand{\Fsp}{\mathcal{F}^{\mathrm{sp}}}
\newcommand{\Fdn}{\mathcal{F}^{\mathrm{dn}}}

\title[Beyond Scalar Rewards: Dense Feedback for LLM Policy Synthesis in Social Dilemmas]{Beyond Scalar Rewards: Dense Feedback for LLM Policy Synthesis in Sequential Social Dilemmas}

\optauthor{%
\Name{Víctor Gallego} \Email{victor.gallego@komorebi.ai}\\
\addr Komorebi AI Technologies, Spain}

\begin{document}

\maketitle

\begin{abstract}%
We propose an LLM harness that generates code-based policy functions for multi-agent environments,
evaluates them with self-play, and refines them using feedback
from previous iterations.  Following the recent line of work in feedback engineering (the design
of which information signals are shown to the LLM during
refinement), we compare sparse feedback (scalar reward only) with
dense feedback (reward plus social metrics: efficiency, equality,
sustainability, peace).
In two Sequential Social Dilemmas (Gathering and
Cleanup) and with two frontier LLMs (Claude Sonnet~4.6, Gemini~3.1~Pro),
dense feedback improves over or matches sparse feedback on all
metrics. We explain this asymmetry via \emph{feedback aliasing}: when the 
scalar reward maps distinct failure modes into the same value (e.g.,
under- vs.\ over-cleaning), social metrics disambiguate and allow the LLM to
diagnose which direction of improvement to take. We conclude that social metrics act as a coordination signal, leading to strategies
such as Voronoi territory partitioning and adaptive cleaner
schedules.

\noindent Code at \url{https://github.com/vicgalle/llm-policies-social-dilemmas}.\end{abstract}


\section{Introduction}
\label{sec:intro}

Sequential Social Dilemmas (SSDs)~\citep{leibo2017multi} are multi-agent
environments where individual rational behavior leads to 
suboptimal collective outcomes, that is, they are the multi-agent analog of the prisoner's dilemma,
extended to Markov games with temporal structure.  Standard multi-agent
reinforcement learning (MARL) struggles with SSDs due to the difficulty of credit assignment, non-stationarity, and vast joint action
spaces~\citep{busoniu2008comprehensive}.

The recent advances in large language models (LLMs) have opened a
different approach: rather than learning policies with gradient-based
optimization in parameter space, the LLM can generate a code-based policy in algorithm space, that is, writing an executable program
that implements complex coordination strategies such as territory division, role assignment, conditional cooperation, and more coordination patterns. This paradigm is related to
FunSearch~\citep{romera2024mathematical} and
Eureka~\citep{ma2024eureka}, and avoids the sample efficiency bottleneck of
MARL because just one LLM generation step can design a sophisticated coordination algorithm that would require millions of RL episodes to discover, thanks to the LLM's prior knowledge and reasoning capabilities.

A natural question then arises when using iterative LLM synthesis: \emph{which
feedback should the LLM receive between iterations?}  We study
\emph{feedback engineering} as a design axis and compare \emph{sparse}
(scalar reward only) against \emph{dense} feedback (reward plus
efficiency, equality, sustainability, and peace).  Across two frontier LLMs
(Claude Sonnet~4.6, Gemini~3.1~Pro) and two canonical SSDs (Gathering,
Cleanup), dense feedback consistently matches or exceeds sparse feedback on
all metrics, with a $54\%$ efficiency gain in Cleanup. In both games and with both LLMs, dense feedback also produces higher equality and sustainability without sacrificing efficiency.
We explain this asymmetry through the lens of \emph{feedback aliasing}: when distinct
failure modes (under- vs.\ over-cleaning) collapse to the same scalar
reward, social metrics break the alias and show the corrective
direction. When no such alias exists (in the Gathering game), both modes converge to
similar performance levels.

\section{Framework}
\label{sec:framework}

\subsection{Sequential Social Dilemmas}
\label{sec:ssds}

An SSD is a partially observable Markov game
$\mathcal{G} = \langle N, \mathcal{S}, \{\mathcal{A}_i\}_{i=1}^N, T, \{R_i\}_{i=1}^N, H \rangle$
with $N$ agents, state space~$\mathcal{S}$ (the gridworld configuration), action spaces~$\mathcal{A}_i$, transition
function~$T$, reward functions~$R_i$, and episode horizon~$H$.  We experiment with two canonical SSDs from the literature. 

\textbf{Gathering}~\citep{leibo2017multi}.
Agents navigate a 2D gridworld and collect apples ($+1$ reward).  Apples
respawn every 25 steps.  Agents can fire a tagging beam
(2~hits remove a rival for 25~steps).  The dilemma is whether agents can coexist
peacefully and share resources, or attack rivals to monopolize
apples (aggression wastes time and reduces total welfare).

\textbf{Cleanup}~\citep{hughes2018inequity}.
A public goods game with two regions: a river that accumulates
waste and an orchard where apples grow.  Apples only regrow when the
river is sufficiently clean.  Agents can fire a cleaning beam
(costs $-1$) to remove waste or collect apples ($+1$).
A penalty beam (costs $-1$, inflicts $-50$ on the target) can tag rivals out for 25~steps.  The dilemma is that cleaning is costly but benefits everyone, and selfish agents take advantage of others' cleaning.

\begin{figure}[h]
\centering
\begin{minipage}{0.6\columnwidth}
  \centering
  \includegraphics[width=\linewidth]{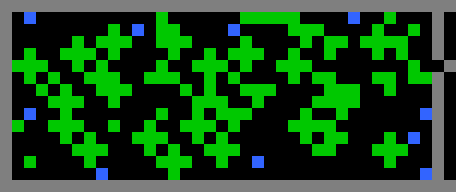}
  \caption*{(a) Gathering}
\end{minipage}\hfill
\begin{minipage}{0.38\columnwidth}
  \centering
  \includegraphics[width=\linewidth]{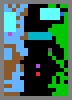}
  \caption*{(b) Cleanup}
\end{minipage}
\caption{Screenshots of the two SSD environments used in our experiments.
\textbf{(a)}~\emph{Gathering}: agents (colored markers) navigate a gridworld to collect apples (green cells); apples respawn on a fixed timer and agents may fire tagging beams to temporarily remove rivals.
\textbf{(b)}~\emph{Cleanup}: agents operate in two regions: a river accumulating waste (brown) and an orchard where apples grow (green); agents must cooperatively clean the river for apples to regrow. Agents may fire beams (cyan) to tag rivals out.}
\label{fig:env-screenshots}
\end{figure}

Both games use 8--9 discrete actions (4 movement directions, 2 rotations,
beam, stand, and clean where appropriate) and episodes of $H\!=\!1000$ steps.
Screenshots of both environments are shown in Figure~\ref{fig:env-screenshots} (Appendix).

Following \citet{perolat2017multi}, we evaluate episode outcomes using four social
metrics.  Let $R_i = \sum_{t=0}^{H-1} r_i^t$ denote agent~$i$'s
episode return.  Then:
\begin{align}
\text{Efficiency:}\quad U &= \frac{1}{H}\textstyle\sum_{i=1}^{N} R_i \label{eq:efficiency}\\
\text{Equality:}\quad E &= 1 - \frac{\sum_{i,j}|R_i - R_j|}{2N\sum_i R_i} \label{eq:equality}\\
\text{Sustainability:}\quad S &= \frac{1}{N}\textstyle\sum_{i=1}^{N} \bar{t}_i \label{eq:sustainability}\\
\text{Peace:}\quad P &= \frac{1}{H}\textstyle\sum_{t=0}^{H-1} \big|\{i : \mathrm{active}_i^t\}\big| \label{eq:peace}
\end{align}
where $\bar{t}_i$ is the mean timestep at which agent~$i$ collects
positive reward (higher means resources remain available later), and
$\mathrm{active}_i^t$ indicates agent~$i$ is not tagged out at step~$t$.

\subsection{Iterative LLM Policy Synthesis}
\label{sec:synthesis}
\begin{figure}[t]
\centering
\begin{minipage}[b]{0.48\textwidth}
  \centering
  \resizebox{\linewidth}{!}{%
  \begin{tikzpicture}[
  phase/.style={
    rectangle, draw=#1!60!black, rounded corners=5pt,
    minimum width=2.9cm, minimum height=1.45cm,
    fill=#1!8, inner sep=5pt, align=center,
    line width=0.6pt
  },
  stepnum/.style={
    circle, fill=#1!55!black, text=white,
    font=\scriptsize\bfseries, inner sep=0pt,
    minimum size=13pt
  },
  ext/.style={
    rectangle, draw=gray!50, rounded corners=3pt, fill=gray!5,
    font=\small, inner sep=4pt, align=center
  },
  fbox/.style={
    rectangle, draw=#1!50!black, rounded corners=3pt, fill=#1!5,
    font=\footnotesize, inner sep=5pt, align=center, minimum height=0.9cm
  },
  annot/.style={font=\footnotesize, align=center, text=black!60},
  arr/.style={-{Stealth[length=6pt,width=5pt]}, semithick, black!60},
  darr/.style={-{Stealth[length=5pt,width=4pt]}, densely dashed, thin, black!35},
]

\node[phase=blue] (synth) at (0,0) {
  \textbf{\small\textsc{Synthesize}}\\[2pt]
  {\footnotesize $\pi_{k+1}\!\gets\!\Mllm(p,\,q_k)$}\\[-1pt]
  {\footnotesize Python policy code}
};
\node[stepnum=blue] at ([shift={(-4pt,4pt)}]synth.north west) {1};

\node[phase=teal] (valid) at (5,0) {
  \textbf{\small\textsc{Validate}}\\[2pt]
  {\footnotesize AST safety check}\\[-1pt]
  {\footnotesize + 50-step smoke test}
};
\node[stepnum=teal] at ([shift={(-4pt,4pt)}]valid.north west) {2};

\node[phase=orange] (eval) at (5,-3.5) {
  \textbf{\small\textsc{Evaluate}}\\[2pt]
  {\footnotesize $N$-agent self-play}\\[-1pt]
  {\footnotesize over $|S|$ seeds}
};
\node[stepnum=orange] at ([shift={(-4pt,4pt)}]eval.north west) {3};

\node[phase=violet] (fb) at (0,-3.5) {
  \textbf{\small\textsc{Feedback}}\\[2pt]
  {\footnotesize Select level $\ell$}\\[-1pt]
  {\footnotesize Construct $q_{k+1}$}
};
\node[stepnum=violet] at ([shift={(-4pt,4pt)}]fb.north west) {4};

\draw[arr] (synth.east) -- node[above,annot] {code $\pi_{k+1}$} (valid.west);
\draw[arr] (valid.south) -- node[right,annot,text width=1.5cm] {verified\\policy} (eval.north);
\draw[arr] (eval.west)  -- node[below,annot] {$\bar{r}_k,\;\mathbf{m}_k$} (fb.east);
\draw[arr] (fb.north)   -- node[left,annot,text width=2cm] {code\,+\,feedback\\prompt $q_{k+1}$} (synth.south);

\draw[-{Stealth[length=5pt,width=4pt]}, densely dotted, semithick, red!45!black]
  (valid.north) .. controls +(0,1.1) and +(0,1.1) ..
  node[above,font=\scriptsize\itshape,text=red!45!black]
  {fail\;$\to$\;retry\;($\le\!R$)}
  (synth.north);

\node[ext, minimum width=1.4cm] (prompt) at (-2.8,1.4)
  {Prompt\;$p$};
\draw[darr] (prompt.south east) -- (synth.north west);

\node[ext, minimum width=1.8cm] (game) at (8,-1.75)
  {\textbf{SSD}\;$\mathcal{G}$};
\draw[darr] (game.west) -- (eval.east);

\node[fbox=blue!60!violet] (sparse) at (-1.3,-5.7) {
  \textsc{Sparse}\;$\Fsp$\\[2pt]
  {\scriptsize $\mathrm{code}(\pi_k),\;\bar{r}_k$}
};
\node[fbox=orange!80!red] (dense) at (2.3,-5.7) {
  \textsc{Dense}\;$\Fdn$\\[2pt]
  {\scriptsize $\mathrm{code}(\pi_k),\;\bar{r}_k,\;\mathbf{m}_k,\;\mathbf{d}$}
};
\draw[darr] (fb.south) -- ++(0,-0.5) -| (sparse.north);
\draw[darr] (fb.south) -- ++(0,-0.5) -| (dense.north);
\node[font=\scriptsize\bfseries, text=gray!45] at (0.5,-5.7) {or};

\node[font=\footnotesize\itshape, text=gray!35] at (2.5,-1.75)
  {iteration $k$};

\begin{pgfonlayer}{background}
  \node[rounded corners=10pt, fill=black!2.5, draw=black!8,
        line width=0.3pt, inner sep=14pt,
        fit=(synth)(valid)(eval)(fb)] {};
\end{pgfonlayer}

\end{tikzpicture}
  }%
\end{minipage}\hfill
\begin{minipage}[b]{0.50\textwidth}
  \centering
  \begin{algorithm}[H]
    \footnotesize                 
    \DontPrintSemicolon
    \caption{Iterative LLM Policy Synthesis}
    \label{alg:synthesis}
    \KwIn{Game $\mathcal{G}$, LLM $\Mllm$, prompt $p$,\\
          iterations $K$, level $\ell$, seeds $S$}
    \KwOut{Final policy $\pi_K$}
    $\pi_0 \gets \Mllm(p,\;\text{``generate initial policy''})$\;\\
    $\mathcal{F}_0^\ell \gets \mathrm{Eval}(\pi_0;\,\mathcal{G},\,S)$\;\\
    \For{$k = 1, \ldots, K$}{
      \For{\textnormal{attempt} $= 1, \ldots, R$}{
        $\pi_k \gets \Mllm\!\big(p,\;q(\pi_{k-1},\mathcal{F}_{k-1}^\ell)\big)$\;
        \lIf{\textnormal{Validate}$(\pi_k)$}{\textbf{break}}
        $q \gets q \,\|\, \text{error\_msg}$\;
      }
      $\mathcal{F}_k^\ell \gets \mathrm{Eval}(\pi_k;\,\mathcal{G},\,S)$\;
    }
    \Return{$\pi_K$}
  \end{algorithm}
\end{minipage}
\caption{\textbf{Iterative LLM policy synthesis.}
\emph{Left:} the four phases cycle. At each iteration~$k$, the LLM
\textsc{Synthesize}s~(\textbf{1}) a Python policy from the system
prompt~$p$ and previous feedback, which is \textsc{Validate}d~(\textbf{2})
via AST checks and a smoke test (retrying on failure up to~$R$ times),
\textsc{Evaluate}d~(\textbf{3}) in $N$-agent self-play, and the results
packaged as either \emph{sparse} or \emph{dense}
\textsc{Feedback}~(\textbf{4}). \emph{Right:} the procedure in pseudocode.}
\label{fig:framework}
\end{figure}

Let $\Pi$ denote the space of code-based policies, i.e., deterministic
functions $\pi: \mathcal{S} \times [N] \to \mathcal{A}$ expressed as Python code.  Each policy has access to the full environment
state and a library of helper functions: breadth-first search (BFS) pathfinding, beam targeting, and basic coordinate transforms. This state access is a deliberate design choice: programmatic policies can be evolved in algorithm space rather than in the reactive observation-to-action space of neural policies. Code as Policies \citep{code-as-policies} demonstrated that LLMs can generate executable robot policy code that processes perception outputs and parameterizes control primitives via few-shot prompting. And Eureka \citep{ma2024eureka} uses LLMs to generate code-based reward functions (rather than policies) from the environment source code. Our work differs from these in that the LLM iteratively synthesizes complete agent policies for a multi-agent setting, where the generated code must simultaneously coordinate across agents sharing the same program.

A frozen LLM~$\Mllm$ acts as a \emph{policy synthesizer}.  Given a system
prompt~$p$ describing the environment API and a feedback
prompt~$q_k$, it generates source code implementing a new policy
$
  \pi_{k+1} = \Mllm\!\left(p,\; q\!\left(\pi_k,\,\mathcal{F}_k^\ell\right)\right),
$
where $\pi_k$ is the previous policy (its source code),
$\mathcal{F}_k^\ell$ is the evaluation feedback at level~$\ell$, and
$q(\cdot)$ builds the user prompt.
All $N$ agents execute the same policy~$\pi_k$ (homogeneous self-play).
The evaluation computes feedback over a set of random seeds~$S$:
\begin{equation*}
  \mathcal{F}_k = \mathrm{Eval}(\underbrace{\pi_k, \ldots,\pi_k}_{N};\;\mathcal{G},\,S)
  = \left(\bar{r}_k,\;\mathbf{m}_k\right)
  \label{eq:eval}
\end{equation*}
where $\bar{r}_k = \frac{1}{N|S|}\sum_{s \in S}\sum_i R_i^{(s)}$ is
the mean per-agent return and $\mathbf{m}_k = (U_k, E_k, S_k, P_k)$ is the
social metrics vector.

Each generated policy undergoes safety checking with a syntax parser
(blocking dangerous operations such as \texttt{eval}, file I/O,
network access, etc.) followed by a smoke test for 50 steps
to catch runtime errors.  If these validations fail, the error message is appended to the prompt and generation is retried (up to 3 attempts).

\subsection{Feedback Engineering}
\label{sec:feedback}

We define two feedback levels~$\ell$ that control what information the LLM
receives between iterations:

\textbf{Sparse feedback} (\textsc{reward-only}).  The LLM only sees the
previous policy's source code and the scalar mean per-agent reward:
\begin{equation}
  \mathcal{F}_k^{\mathrm{sp}} = \left(\,\mathrm{code}(\pi_k),\;\bar{r}_k\,\right)
  \label{eq:sparse}
\end{equation}

\textbf{Dense feedback} (\textsc{reward+social}).  The LLM additionally
receives the full social metrics vector together with natural language
definitions of each metric:
\begin{equation}
  \mathcal{F}_k^{\mathrm{dn}} = \left(\,\mathrm{code}(\pi_k),\;\bar{r}_k,\;\mathbf{m}_k,\;\mathbf{d}\,\right)
  \label{eq:dense}
\end{equation}
where $\mathbf{d}$ contains textual definitions
(e.g., ``\emph{Equality: fairness of reward distribution, 1.0 =
perfectly equal}''). We avoid leaking environment information in these definitions to ensure a fair comparison between methods.

In both modes the system prompt instructs the LLM to
\emph{maximize per-agent reward}: the social metrics in dense feedback are presented as informational context instead of being displayed as optimization targets. Both modes use a neutral framing ``all agents run the same
code'' (no adversarial language, nor placing emphasis on cooperation).
At iteration~0 the LLM generates a policy from scratch; the following
iterations receive the previous policy's code together with its feedback.
The crucial design question is whether~$\ell = \mathrm{sp}$
or~$\ell = \mathrm{dn}$ produces better policies.


\section{Experiments}
\label{sec:experiments}

\subsection{Setup}

We run both SSDs with $N\!=\!10$ agents on large map variants,
$K\!=\!3$ refinement iterations, $|S|\!=\!5$ evaluation seeds, and
3~independent runs per configuration.
We evaluate two frontier LLMs, \textbf{Claude Sonnet~4.6} and
\textbf{Gemini~3.1~Pro}, both with maximum thinking budget.
For each model and game combination, we compare three settings:
\textsc{zero-shot} (no refinement), \textsc{reward-only} (sparse
feedback), and \textsc{reward+social} (dense feedback).
As other baselines, we run a tabular \textbf{Q-learner} with
hand-crafted features and cooperative reward shaping, a hand-coded
\textbf{BFS Collector} (nearest-apple navigation, no beaming), and
\textbf{GEPA}~\citep{gepa}, an LLM-driven prompt optimizer using the same
Gemini~3.1~Pro and matched compute budget; GEPA's reflection step receives
only scalar reward, similar to \textsc{reward-only}'s information.
Full details are in Appendix~\ref{app:setup}.

\subsection{Main Results}

\begin{table*}[t]
\centering
\caption{Results across two SSDs, two LLMs, and three feedback configurations.
LLM values show the mean over $3 \times 5$ independent runs (min--max in parentheses).
$U$: efficiency (Eq.~\ref{eq:efficiency}).
$E$: equality (Eq.~\ref{eq:equality}).
$S$: sustainability (Eq.~\ref{eq:sustainability}).
Bold marks the best value per game$\,\times\,$model block.
Baselines (bottom of each game block) are non-LLM methods for reference.}
\label{tab:main}
\footnotesize
\begin{tabular}{@{}ll l ccc @{}}
\toprule
\textbf{Game} & \textbf{Model} & \textbf{Feedback} & $U$ & $E$ & $S$  \\
\midrule
\multirow{6}{*}{Gathering}
  & \multirow{3}{*}{Claude Sonnet 4.6}
  & \textsc{zero-shot}                         & 1.85\,{\scriptsize(1.52--2.02)}  & 0.52\,{\scriptsize(0.35--0.63)}  & 298.6\,{\scriptsize(180--374)}   \\
  & & \textsc{reward-only}           & 3.47\,{\scriptsize(3.39--3.56)}  & 0.72\,{\scriptsize(0.61--0.87)}  & 402.9\,{\scriptsize(314--500)}   \\
  & & \textsc{reward+social}       & \textbf{3.53}\,{\scriptsize(1.60--4.58)}  & \textbf{0.84}\,{\scriptsize(0.63--0.94)}  & \textbf{452.7}\,{\scriptsize(356--502)}   \\
  \cmidrule{2-6}
  & \multirow{3}{*}{Gemini 3.1 Pro}
  & \textsc{zero-shot}                         & 3.71\,{\scriptsize(1.90--4.46)}  & 0.79\,{\scriptsize(0.35--0.96)}  & 443.2\,{\scriptsize(180--504)}   \\
  & & \textsc{reward-only}           & 4.58\,{\scriptsize(4.48--4.63)}  & \textbf{0.97}\,{\scriptsize(0.97--0.97)}  & 502.5\,{\scriptsize(502--503)}   \\
  & & \textsc{reward+social}       & \textbf{4.59}\,{\scriptsize(4.50--4.65)}  & \textbf{0.97}\,{\scriptsize(0.97--0.97)}  & \textbf{502.7}\,{\scriptsize(501--505)}   \\
  \cmidrule{2-6}
  & \multicolumn{2}{l}{\color{gray}\emph{GEPA (Gemini 3.1 Pro)}}
                                    & \color{gray}3.45\,{\scriptsize(3.16--3.95)}  & \color{gray}0.91\,{\scriptsize(0.83--0.96)}  & \color{gray}496.2\,{\scriptsize(491--499)}   \\
  & \multicolumn{2}{l}{\color{gray}\emph{Q-learner}}
                                    & \color{gray}0.77\,{\scriptsize(0.73--0.81)}  & \color{gray}0.83\,{\scriptsize(0.81--0.85)}  & \color{gray}508.2\,{\scriptsize(504--513)}   \\
  & \multicolumn{2}{l}{\color{gray}\emph{BFS Collector}}
                                    & \color{gray}1.29  & \color{gray}0.54  & \color{gray}489.5   \\
\midrule
\multirow{6}{*}{Cleanup}
  & \multirow{3}{*}{Claude Sonnet 4.6}
  & \textsc{zero-shot}                          & $-$1.01\,{\scriptsize($-$7.90--0.93)}  & $-$3.06\,{\scriptsize($-$15.9--1.00)}  & 137.0\,{\scriptsize(19--214)}   \\
  & & \textsc{reward-only}           & 1.14\,{\scriptsize(1.04--1.32)}  & $-$0.47\,{\scriptsize($-$0.66--$-$0.11)}  & 233.0\,{\scriptsize(225--245)}  \\
  & & \textsc{reward+social}       & \textbf{1.37}\,{\scriptsize(1.16--1.73)}  & \textbf{0.09}\,{\scriptsize($-$0.15--0.55)}  & \textbf{294.6}\,{\scriptsize(209--456)}  \\
  \cmidrule{2-6}
  & \multirow{3}{*}{Gemini 3.1 Pro}
  & \textsc{zero-shot}                          & 0.45\,{\scriptsize($-$2.32--1.28)}  & $-$0.45\,{\scriptsize($-$1.38--0.73)}  & 274.1\,{\scriptsize(158--421)}  \\
  & & \textsc{reward-only}           & 1.79\,{\scriptsize(1.16--2.57)}  & 0.13\,{\scriptsize($-$0.47--0.47)}  & 386.0\,{\scriptsize(186--503)}  \\
  & & \textsc{reward+social}       & \textbf{2.75}\,{\scriptsize(1.44--3.49)}  & \textbf{0.54}\,{\scriptsize(0.33--0.67)}  & \textbf{432.6}\,{\scriptsize(297--501)}  \\
  \cmidrule{2-6}
  & \multicolumn{2}{l}{\color{gray}\emph{GEPA (Gemini 3.1 Pro)}}
                                    & \color{gray}0.77\,{\scriptsize(0.58--1.07)}  & \color{gray}$-$1.75\,{\scriptsize($-$2.54--$-$0.62)}  & \color{gray}209.5\,{\scriptsize(184--260)}  \\
  & \multicolumn{2}{l}{\color{gray}\emph{Q-learner}}
                                    & \color{gray}$-$0.16\,{\scriptsize($-$0.28--$-$0.06)}  & \color{gray}0.20\,{\scriptsize($-$0.37--1.00)}  & \color{gray}208.6\,{\scriptsize(121--301)}  \\
  & \multicolumn{2}{l}{\color{gray}\emph{BFS Collector}}
                                    & \color{gray}0.10  & \color{gray}0.61  & \color{gray}16.4   \\
\bottomrule
\end{tabular}
\end{table*}

Table~\ref{tab:main} presents results across both games, both models,
and all feedback configurations.  Three findings are of special interest.

\paragraph{Finding 1: LLM policy synthesis dominates traditional and
prompt-level baselines.}
All refined LLM policies dramatically outperform non-LLM baselines.  In
Gathering, the best configuration (Gemini, dense, $U\!=\!4.59$) achieves
$6.0\times$ the Q-learner ($U\!=\!0.77$) and $3.6\times$ the BFS heuristic
($U\!=\!1.29$); in Cleanup, $U\!=\!2.75$ vs.\ $-0.16$, as tabular Q-learning
fails at the cleaning--harvesting credit assignment.  Iterative refinement
is important: Claude's \textsc{zero-shot} scores $U\!=\!{-}1.01$ in Cleanup,
rising to $U\!=\!1.14$--$1.37$ after 3 iterations.  GEPA, which optimizes
the system prompt rather than the code, trails direct code-level iteration
by $25\%$ in Gathering ($U\!=\!3.45$ vs.\ $4.59$) and $3.6\times$ in
Cleanup ($U\!=\!0.77$ vs.\ $2.75$), but with poor, negative equality
($E\!=\!{-}1.75$) indicating free-riding: code-level feedback is
substantially more effective than prompt-level meta-optimization for
cooperative strategy discovery.

\paragraph{Finding 2: Dense feedback consistently matches or exceeds
sparse feedback.}
Across all four game and model combinations,
\textsc{reward+social} achieves equal or higher efficiency than
\textsc{reward-only}.  The advantage is most pronounced in Cleanup:
Gemini gains $54\%$ ($U$: $2.75$ vs.\ $1.79$) and Claude $20\%$
($1.37$ vs.\ $1.14$).  In Gathering the modes perform similarly,
with dense feedback holding a slight edge for Claude
($3.53$ vs.\ $3.47$).

\paragraph{Finding 3: Social metrics serve as a coordination signal.}
Dense feedback simultaneously improves the social metrics without tradeoffs (e.g., Cleanup/Gemini: $E$ rises from
$0.13$ to $0.54$, $S$ from $386$ to $433$ while $U$ also peaks).
We inspect the generated code (Appendix~\ref{app:code}) to reveal how:
under dense feedback, the LLM writes adaptive cleaner
schedules that scale the cleaning team with pollution level and
BFS-Voronoi territory partitioning with zero aggression in
Gathering. Sparse feedback policies instead use fixed cleaning roles
and multi-tier combat systems that waste actions.
Between models, Gemini~3.1~Pro is consistently stronger and achieves lower variance
than Claude Sonnet~4.6 (e.g., Gathering/dense: $U\!\in\![4.50,4.65]$ vs.\
$[1.60,4.58]$), with the largest gap in Cleanup ($2.75$ vs.\ $1.37$).

\subsection{Why Dense Feedback Helps: Feedback Aliasing}
\label{sec:aliasing}

The asymmetry in Table~\ref{tab:main} (dense feedback's efficiency gain is
large in Cleanup but negligible in Gathering) admits an intuitive explanation
we call \emph{feedback aliasing}: whether the scalar reward alone is just enough information to identify the correct refinement direction.
In Cleanup, total reward is a concave function of the number of cleaners
$n_c$ with an interior optimum.  Below this peak, \emph{two} distinct
failure modes produce the same scalar reward but require opposite
corrections: under-cleaning (too few cleaners, apples starved by
pollution) and over-cleaning (too many cleaners, cost exceeds marginal
benefit).  Social metrics break this alias: under-cleaning leads to
low sustainability, over-cleaning to low equality (cleaners bear the cost while harvesters capture the reward).  Dense feedback thus gives the LLM a
diagnostic signal that is lacking in sparse feedback, and this is visible in the
code: under dense feedback the LLM writes adaptive schedules that
scale $n_c$ with pollution (Appendix~\ref{app:cleanup}), while sparse
feedback produces fixed allocations that cannot self-correct.
Gathering has no such alias since the coordination problem reduces to
reducing territorial overlap on a single axis. Hence, the scalar reward already indicates the corrective direction, and both modes converge.

\section{Related Work}
\label{sec:related}

\paragraph{Sequential Social Dilemmas.}
SSDs were introduced by \citet{leibo2017multi} (Gathering) and extended
to public goods settings by \citet{hughes2018inequity} (Cleanup);
\citet{perolat2017multi} formalized the social outcome metrics we use in the experiments.

\paragraph{LLMs for policy and program synthesis.}
FunSearch~\citep{romera2024mathematical} evolves programs for
combinatorial problems; Eureka~\citep{ma2024eureka} synthesizes reward
functions; Voyager~\citep{wang2024voyager} and Code as
Policies~\citep{code-as-policies} generate executable skill code for
single agent control; ReEvo~\citep{ye2024reevo} evolves heuristics via
reflection.  We target \emph{multi-agent} policy code where a single
program must coordinate across agents.

\paragraph{LLM reflection and feedback.}
Reflexion~\citep{shinn2023reflexion},
Self-Refine~\citep{madaan2023selfrefine}, OPRO~\citep{yang2024large}, and
GEPA~\citep{gepa} all demonstrate that structured verbal feedback loops
improve LLM outputs; ERL~\citep{shi2026experiential} internalizes such
reflection via self-distillation.  We complement this line by varying the
content of feedback (scalar vs.\ multi-objective social metrics)
in a multi-agent setting.

\noindent An extended version of this section is located at Appendix \ref{app:extended_related}.

\section{Discussion and Conclusion}
\label{sec:conclusion}

We conclude that richer feedback helps: dense social metrics consistently match or exceed
sparse scalar reward across two games and two frontier LLMs.  Rather than
triggering an excessive optimization of fairness, social metrics act as a
coordination signal that disambiguates distinct failure modes collapsed by the
scalar reward level (the feedback aliasing argument of Section~\ref{sec:aliasing}).
This work extends the LLM-reflection literature~\citep{shinn2023reflexion,
madaan2023selfrefine} to multi-agent settings where the feedback
dimensions capture social outcomes.

\paragraph{Limitations and future work.}
Our SSDs are small-scale; scaling to larger environments, heterogeneous
per-agent policies, and intermediate feedback levels (e.g., partial social
metrics) are natural next steps.  On the safety side, an open question is
whether exploits emerge organically during iterative synthesis without
explicit adversarial prompting. While in the previous experiments we didn't find evidence of the LLM reward hacking the environment, we separately examined whether LLMs can reward-hack our framework by
mutating the environment object they receive. An adversarially prompted Claude Opus~4.6 autonomously discovered five distinct attacks
(Appendix~\ref{app:adversarial}), with attacks that bypass environment dynamics and
amplify reward up to $59\times$ while simultaneously improving measured
social metrics, a concern with direct implications for
verification pipelines in LLM policy synthesis.

\bibliography{sample}
\clearpage
\appendix

\section{Environment Details}
\label{app:env-screenshots}

\textbf{Gathering}~\citep{leibo2017multi}.
Agents navigate a 2D gridworld and collect apples ($+1$ reward).  Apples
respawn on a fixed 25-step timer.  Agents may fire a tagging beam
(2~hits remove a rival for 25~steps).  The dilemma: agents can coexist
peacefully and share resources, or attack rivals to monopolize
apples, but aggression wastes time and reduces total welfare.

\textbf{Cleanup}~\citep{hughes2018inequity}.
A public goods game with two regions: a river that accumulates
waste, and an orchard where apples grow.  Apples only regrow when the
river is sufficiently clean.  Agents can fire a cleaning beam
(costs $-1$) to remove waste, or collect apples ($+1$).
A penalty beam (costs $-1$, inflicts $-50$ on the target) can tag
rivals out for 25~steps.  The dilemma: cleaning is costly but benefits
everyone; purely selfish agents free-ride on others' cleaning.

Both games use 8--9 discrete actions (4 movement directions, 2 rotations,
beam, stand, and optionally clean) and episodes of $H\!=\!1000$ steps.
Gathering uses a $38\!\times\!16$ gridworld with ${\sim}120$ apple spawns;
Cleanup uses a gridworld with separate river and orchard regions.

\section{Baselines and Experimental Setup}
\label{app:setup}

\paragraph{Q-learner.}
Tabular Q-learning with a shared Q-table and non-trivial feature
engineering: 7~hand-crafted discrete features in Gathering
(BFS direction and distance to nearest apple, local apple density,
nearest-agent direction and distance, beam-path check, own hit count;
4\,320 states) and 8~features in Cleanup (adding BFS to nearest waste,
global waste density, and a can-clean check; 11\,664 states), plus
cooperative reward shaping
($0.5 \cdot r_i + 0.5 \cdot \bar{r} - \text{beam penalty}$, with an
additional cleaning bonus in Cleanup).  Trained for 1000 episodes with
$\varepsilon$-greedy exploration.

\paragraph{BFS Collector.}
Hand-coded heuristic: BFS to the nearest apple, never beams or cleans.

\paragraph{GEPA.}
Genetic-Pareto Prompt Optimization~\citep{gepa}, an LLM-based
meta-optimizer that iteratively refines the \emph{system prompt}
(not the policy code) using a reflection LM.  We run GEPA with
Gemini~3.1~Pro for both generation and reflection, with $K\!=\!3$
reflection iterations and $n_{\text{eval}}\!=\!5$ evaluation seeds per
candidate, matching the compute budget of the iterative code-level
methods.  GEPA's reflection LM receives only scalar reward; social
metric definitions are not included, to avoid information leakage.

\section{Generated Policy Analysis}
\label{app:code}

All code excerpts below are verbatim LLM output,
extracted from the best-performing iteration of representative runs.
Comments in \texttt{[brackets]} are ours.

\subsection{Gathering: Territory Strategies}
\label{app:gathering}

Under dense feedback, the LLM discovers \emph{BFS-Voronoi territory
partitioning}: a multi-source flood-fill simultaneously from all alive
agents computes true shortest-path ownership of every cell, correctly
handling walls where Manhattan distance fails.  The policy is purely
cooperative---no agent ever fires the tagging beam.

\begin{lstlisting}[style=pysmall, caption={Gathering --- dense feedback (BFS-Voronoi, zero aggression).}]
# [Multi-source BFS: O(H*W) flood-fill assigns territory]
bfs_q = deque()
dist_map = {}

for i in range(env.n_agents):
    if int(env.agent_timeout[i]) > 0:
        continue
    r, c = int(env.agent_pos[i][0]), int(env.agent_pos[i][1])
    dist_map[(r, c)] = (0, i)
    bfs_q.append((r, c, 0, i))

while bfs_q:
    r, c, d, owner = bfs_q.popleft()
    cur = dist_map.get((r, c), (10**9, 10**9))
    if d > cur[0] or (d == cur[0] and owner > cur[1]):
        continue
    for dr2, dc2 in ((-1,0),(1,0),(0,-1),(0,1)):
        nr, nc = r + dr2, c + dc2
        if 0<=nr<H and 0<=nc<W and not walls[nr][nc]:
            nd = d + 1
            prev2 = dist_map.get((nr,nc), (10**9, 10**9))
            if nd<prev2[0] or (nd==prev2[0] and owner<prev2[1]):
                dist_map[(nr,nc)] = (nd, owner)
                bfs_q.append((nr, nc, nd, owner))

# Phase 1: BFS-navigate to nearest owned alive apple
# Phase 2: camp at nearest owned dead spawn (wait for respawn)
# Phase 3: global fallback to any reachable alive apple
\end{lstlisting}

Under sparse feedback, the LLM discovers \emph{column-strip territory}
(a simpler $O(1)$ assignment) but also develops a multi-tier combat system
that wastes actions on beaming and chasing:

\begin{lstlisting}[style=pysmall, caption={Gathering --- sparse feedback (column strips + combat tiers).}]
# [Static column-strip territory: O(1) assignment]
zone_width = env.width / env.n_agents  # 38/10 = 3.8
zone_start = int(agent_id * zone_width)
zone_end = int((agent_id+1) * zone_width)

# [Multi-tier combat system wastes actions]
# Tier 1: Kill shot directly ahead -- always fire
# Tier 2: Rotate once to land kill shot
# Tier 3: Chase wounded opponents within range 8
# Tier 4: First hit on very close targets (range <= 2)

# [Evasion when one hit from being tagged]
if my_hits >= hits_to_tag - 1 and threats:
    # Flee away from nearest threat
    ...

# [Territory-based collection: home zone first, global fallback]
home_apples = {pos for pos in alive_apples
               if zone_start <= pos[1] < zone_end}
result = bfs_to_target_set(env, agent_id, home_apples)
\end{lstlisting}

The BFS-Voronoi policy achieves higher reward than column strips
because territory adapts dynamically as agents move, and the absence
of combat means every action is spent collecting.

\subsection{Cleanup: Cleaner Allocation Strategies}
\label{app:cleanup}

The most impactful difference between dense and sparse policies is in
\emph{cleaner allocation}---how many agents are assigned to the costly
but socially necessary cleaning role---and \emph{cleaning efficiency}---how
effectively each cleaner removes waste.

Under dense feedback, the LLM develops a waste-adaptive scaling schedule
combined with optimized beam positioning:

\begin{lstlisting}[style=pysmall, caption={Cleanup --- dense feedback (adaptive scaling + optimal beam positioning).}]
# [Waste-adaptive schedule: up to 7/10 agents clean]
if   waste_ratio >= 0.8: n_cleaners = 7
elif waste_ratio >= 0.6: n_cleaners = 5
elif waste_ratio >= 0.4: n_cleaners = 3
elif waste_ratio >= 0.2: n_cleaners = 2
elif waste_ratio >= 0.07: n_cleaners = 1
else:                     n_cleaners = 0

# [Stable ID-based assignment: lowest IDs are permanent cleaners]
is_cleaner = agent_id < n_cleaners

# [Smart cleaning: search 9x9 area around waste centroid
#  for (row, col, orientation) maximizing waste in beam path]
cr, cc = int(np.mean(wr)), int(np.mean(wc))
for dr in range(-4, 5):
    for dc in range(-4, 5):
        r2, c2 = cr + dr, cc + dc
        if not env.walls[r2, c2]:
            for o in range(4):
                cnt = beam_count_at(r2, c2, o)
                if cnt > best_count:
                    best_count = cnt
                    best_pos = (r2, c2, o)
\end{lstlisting}\label{listing:3}

Under sparse feedback, the LLM assigns fixed cleaning roles to
specific agents based on hard-coded thresholds, with simpler beam
targeting:

\begin{lstlisting}[style=pysmall, caption={Cleanup --- sparse feedback (fixed agent-specific thresholds).}]
# [Only 4 of 10 agents can ever clean; hard-coded per-agent]
THRESHOLDS = {0: 0.15, 5: 0.20, 1: 0.40, 6: 0.45}
my_threshold = THRESHOLDS.get(agent_id, 2.0)  # 2.0 = never
is_cleaner = waste_fraction > my_threshold

# [Simple cleaning: best orientation from current position only]
counts = [count_waste_in_dir(o) for o in range(4)]
best_dir = counts.index(max(counts))
if best_dir == cur_orient:
    return CLEAN
\end{lstlisting}

The dense-feedback policy is more effective for two reasons:
(1)~it allocates \emph{more cleaners when pollution is high}, preventing
the ecosystem collapse that occurs when waste overwhelms a small fixed
cleaning force; and (2)~it \emph{navigates to optimal firing positions}
rather than cleaning only from the current location, dramatically
increasing waste removal per action.

\section{Prompts}
\label{app:prompts}

This appendix reproduces the exact prompts used in all experiments.
Each LLM call consists of a \emph{system prompt} (fixed per game) and a
\emph{user prompt} (constructed per iteration).  The system prompt
defines the environment API, helper functions, and output format.  The
user prompt provides the iteration context: previous policy code and
performance feedback.

The two feedback modes differ only in the user prompt at iterations
$k \geq 1$:
\begin{itemize}
\item \textsc{reward-only} (sparse): shows the previous policy source
      and scalar mean per-agent reward.
\item \textsc{reward+social} (dense): additionally includes social
      metric definitions and values (efficiency, equality,
      sustainability, peace).
\end{itemize}
At iteration~0, both modes use identical prompts since no history
exists.


\subsection{Gathering: System Prompt}
\label{app:prompt-gathering-sys}

The following system prompt is used for all Gathering experiments
(both \textsc{reward-only} and \textsc{reward+social}).

\begin{lstlisting}[style=pysmall, caption={Gathering system prompt (verbatim).}, basicstyle=\ttfamily\tiny]
You are an expert game-theoretic AI designing policies for a multi-agent
Sequential Social Dilemma (the Gathering game).

## Environment Summary

- 2D gridworld. Agents collect apples (+1 reward each). Apples respawn after
  25 steps. Agents can fire a "tagging beam" that temporarily removes rivals
  for 25 steps (requires 2 hits to tag in Gathering).
- Episode length: 1000 steps.
- 8 actions: FORWARD(0), BACKWARD(1), STEP_LEFT(2), STEP_RIGHT(3),
  ROTATE_LEFT(4), ROTATE_RIGHT(5), BEAM(6), STAND(7)
- Agents move in 4 cardinal directions WITHOUT needing to rotate first
  (strafe movement). Rotation only matters for the beam direction.

## Environment API (available in your policy's namespace)

```python
# env attributes you can read:
env.agent_pos        # np.array shape (n_agents, 2) -- [row, col] per agent
env.agent_orient     # np.array shape (n_agents,) -- 0=N, 1=E, 2=S, 3=W
env.agent_timeout    # np.array shape (n_agents,) -- >0 means agent is removed
env.agent_beam_hits  # np.array shape (n_agents,) -- hits accumulated toward tag
env.apple_alive      # np.array shape (n_apples,) bool -- which apples exist
env._apple_pos       # np.array shape (n_apples, 2) -- [row, col] per apple spawn
env.walls            # np.array shape (H, W) bool -- wall map
env.height, env.width                  # map dimensions
env.n_agents, env.n_apples             # counts
env.beam_length, env.beam_width        # beam parameters (20, 1)
env.hits_to_tag, env.timeout_steps     # 2 hits to tag, 25 step timeout
```

## Helper functions available in your namespace

```python
from gathering_env import Action, Orientation, _ROTATIONS, NUM_ACTIONS

bfs_nearest_apple(env, agent_id) -> Optional[Tuple[int,int]]
bfs_to_target_set(env, agent_id, target_set) -> Optional[Tuple[int,int]]
bfs_toward(env, agent_id, target_r, target_c) -> Optional[Tuple[int,int]]
direction_to_action(dr, dc, orientation) -> int
get_opponents(env, agent_id) -> list
_beam_targets_for_orient(env, ar, ac, orient_val, opponents) -> list
_rotation_distance(cur, target) -> int
greedy_action(env, agent_id) -> int
exploitative_action(env, agent_id) -> int
# Also available: np (numpy), deque (from collections)
```

## Your task

Write a Python function called `policy` with this exact signature:

```python
def policy(env, agent_id) -> int:
    """Return an action (int 0-7) for the given agent."""
    ...
```

The function must:
1. Return an integer 0-7 (an Action value)
2. Be deterministic given the environment state
3. Only use the env attributes and helper functions listed above
4. Not import any modules (numpy and deque are pre-loaded)
5. Not use eval(), exec(), open(), or __import__

## Working Example (seed BFS policy)

```python
def policy(env, agent_id) -> int:
    """BFS greedy: go to nearest apple, never beam."""
    if int(env.agent_timeout[agent_id]) > 0:
        return 7  # STAND while removed
    result = bfs_nearest_apple(env, agent_id)
    if result is None:
        return 7  # No reachable apple -- stand
    dr, dc = result
    return direction_to_action(dr, dc, int(env.agent_orient[agent_id]))
```

IMPORTANT:
- Always check `if result is None` before unpacking BFS results.
- Always cast env arrays to int when comparing.
- Always return a plain int (0-7), never a tuple or None.
- Put your code in a single ```python ... ``` block.
- Before the code block, explain your reasoning for the policy design.
\end{lstlisting}


\subsection{Cleanup: System Prompt}
\label{app:prompt-cleanup-sys}

The following system prompt is used for all Cleanup experiments.

\begin{lstlisting}[style=pysmall, caption={Cleanup system prompt (verbatim).}, basicstyle=\ttfamily\tiny]
You are an expert game-theoretic AI designing policies for a multi-agent
Sequential Social Dilemma (the Cleanup game).

## Environment Summary

- 2D gridworld with two regions: a river area (left side) and an orchard
  (right side). A stream separates the two regions.
- Agents collect apples in the orchard (+1 reward each).
- Waste (pollution) accumulates in the river over time.
- Episode length: 1000 steps.
- 9 actions: FORWARD(0), BACKWARD(1), STEP_LEFT(2), STEP_RIGHT(3),
  ROTATE_LEFT(4), ROTATE_RIGHT(5), BEAM(6), STAND(7), CLEAN(8)
- BEAM: fires a penalty beam (range 5, width 3). Costs -1 reward to fire.
  Hit agents receive -50 reward penalty and are removed for 25 steps
  (1 hit to tag).
- CLEAN: fires a cleaning beam (range 5, width 3). Costs -1 reward to fire.
  Removes waste cells in the beam's path, restoring clean river.
- Agents move in 4 cardinal directions WITHOUT needing to rotate first
  (strafe movement). Rotation only matters for the beam/clean direction.

## Environment API (available in your policy's namespace)

```python
# env attributes you can read:
env.agent_pos        # np.array shape (n_agents, 2) -- [row, col] per agent
env.agent_orient     # np.array shape (n_agents,) -- 0=N, 1=E, 2=S, 3=W
env.agent_timeout    # np.array shape (n_agents,) -- >0 means agent is removed
env.agent_beam_hits  # np.array shape (n_agents,) -- hits accumulated toward tag
env.apple_alive      # np.array shape (n_apples,) bool -- which apples exist
env._apple_pos       # np.array shape (n_apples, 2) -- [row, col] per apple spawn
env.walls            # np.array shape (H, W) bool -- wall map
env.waste            # np.array shape (H, W) bool -- True where waste exists
env.river_cells_set  # set of (row, col) -- all river cell positions
env.stream_cells_set # set of (row, col) -- stream cell positions
env.height, env.width                  # map dimensions
env.n_agents, env.n_apples             # counts
env.beam_length, env.beam_width        # beam/clean parameters (5, 3)
env.hits_to_tag, env.timeout_steps     # 1 hit to tag, 25 step timeout
```

## Helper functions available in your namespace

```python
from cleanup_env import CleanupAction, NUM_CLEANUP_ACTIONS
from gathering_env import Orientation, _ROTATIONS

bfs_nearest_apple(env, agent_id) -> Optional[Tuple[int,int]]
bfs_to_target_set(env, agent_id, target_set) -> Optional[Tuple[int,int]]
bfs_toward(env, agent_id, target_r, target_c) -> Optional[Tuple[int,int]]
direction_to_action(dr, dc, orientation) -> int
get_opponents(env, agent_id) -> list
_beam_targets_for_orient(env, ar, ac, orient_val, opponents) -> list
_rotation_distance(cur, target) -> int
greedy_action(env, agent_id) -> int
# Also available: np (numpy), deque (from collections)
```

## Your task

Write a Python function called `policy` with this exact signature:

```python
def policy(env, agent_id) -> int:
    """Return an action (int 0-8) for the given agent."""
    ...
```

The function must:
1. Return an integer 0-8 (a CleanupAction value)
2. Be deterministic given the environment state
3. Only use the env attributes and helper functions listed above
4. Not import any modules (numpy and deque are pre-loaded)
5. Not use eval(), exec(), open(), or __import__

## Working Example (seed BFS policy)

```python
def policy(env, agent_id) -> int:
    """BFS greedy: go to nearest apple, never beam or clean."""
    if int(env.agent_timeout[agent_id]) > 0:
        return 7  # STAND while removed
    result = bfs_nearest_apple(env, agent_id)
    if result is None:
        return 7  # No reachable apple -- stand
    dr, dc = result
    return direction_to_action(dr, dc, int(env.agent_orient[agent_id]))
```

IMPORTANT:
- Always check `if result is None` before unpacking BFS results.
- Always cast env arrays to int when comparing.
- Always return a plain int (0-8), never a tuple or None.
- Put your code in a single ```python ... ``` block.
- Before the code block, explain your reasoning for the policy design.
\end{lstlisting}


\subsection{User Prompts}
\label{app:prompt-user}

At each iteration $k$, a user prompt is constructed from the current
state.  Below we show the two templates.

\subsubsection{Iteration 0 (both modes)}

When no prior policy exists, the user prompt is identical for both
\textsc{reward-only} and \textsc{reward+social}. This is the \textsc{zero-shot} variant from the experiments:

\begin{lstlisting}[style=pysmall, caption={User prompt at iteration~0 (both modes).}, basicstyle=\ttfamily\tiny]
## Iteration 0/K: Write the initial policy

No prior policy exists yet. All agents will run the same code.
Your task is to write a first policy that maximizes per-agent reward.

## Instructions

Write a policy that maximizes per-agent reward. All agents will run your
exact same code simultaneously. There are {N} agents on a {W}x{H} map
with ~{A} apple spawns.
{env_hint}

Write your `policy(env, agent_id) -> int` function (returns 0-{max_action}).
\end{lstlisting}

\noindent where \texttt{\{env\_hint\}} is game-specific:
\begin{itemize}
\item \textbf{Gathering}: \emph{``Apples respawn every 25 steps. It takes 2
  beam hits to tag out an agent.''}
\item \textbf{Cleanup}: \emph{``Waste accumulates in the river over time.
  BEAM costs $-1$ to fire ($-50$ to target, 1 hit tags out for 25 steps).
  CLEAN costs $-1$ to fire (removes waste in beam path).''}
\end{itemize}

\subsubsection{Iteration $k \geq 1$: \textsc{reward-only} (sparse)}

The previous policy's source code and scalar reward are shown:

\begin{lstlisting}[style=pysmall, caption={User prompt at iteration $k \geq 1$, \textsc{reward-only} mode.}, basicstyle=\ttfamily\tiny]
## Iteration k/K: Write an improved policy

The following policy is currently used by all agents. All agents run the
same code. Your task is to write an improved version that maximizes
per-agent reward.

### Current policy: **P{k-1}_reward**

```python
{previous_policy_source_code}
```

## Results from previous iterations

- Iteration 0: Avg agent reward=X.X
- Iteration 1: Avg agent reward=Y.Y
...

## Instructions

Write a policy that maximizes per-agent reward. All agents will run your
exact same code simultaneously. There are {N} agents on a {W}x{H} map
with ~{A} apple spawns.
{env_hint}

Write your `policy(env, agent_id) -> int` function (returns 0-{max_action}).
\end{lstlisting}

\subsubsection{Iteration $k \geq 1$: \textsc{reward+social} (dense)}

In addition to the scalar reward, social metric definitions and values
are included:

\begin{lstlisting}[style=pysmall, caption={User prompt at iteration $k \geq 1$, \textsc{reward+social} mode.}, basicstyle=\ttfamily\tiny]
## Iteration k/K: Write an improved policy

The following policy is currently used by all agents. All agents run the
same code. Your task is to write an improved version that maximizes
per-agent reward.

### Current policy: **P{k-1}_rall**

```python
{previous_policy_source_code}
```

## Results from previous iterations

### Social Metrics (definitions)

- **Efficiency**: collective apple collection rate across all agents
  (higher = more apples collected per step).
- **Equality**: fairness of reward distribution between agents
  (1.0 = perfectly equal, negative = highly unequal).
- **Sustainability**: long-term apple availability -- measures whether
  resources are preserved over the episode (higher = apples remain
  available later in the episode).
- **Peace**: absence of aggressive beaming -- counts agents not involved
  in attack beam conflicts (higher = less aggression). Using the CLEAN
  beam to remove waste does NOT reduce peace.

- Iteration 0: Avg agent reward=X.X | efficiency=A.AAA,
  equality=B.BBB, sustainability=C.C, peace=D.D
- Iteration 1: Avg agent reward=Y.Y | efficiency=...
...

## Instructions

Write a policy that maximizes per-agent reward. All agents will run your
exact same code simultaneously. There are {N} agents on a {W}x{H} map
with ~{A} apple spawns.
{env_hint}

Write your `policy(env, agent_id) -> int` function (returns 0-{max_action}).
\end{lstlisting}


\section{Related Work (extended)}
\label{app:extended_related}
\paragraph{Sequential Social Dilemmas.}
\citet{leibo2017multi} introduced SSDs as temporally extended Markov games
exhibiting cooperation--defection tension, instantiated in the Gathering
gridworld.  \citet{hughes2018inequity} proposed the Cleanup game as a
public goods variant requiring costly pro-social labor.
\citet{perolat2017multi} formalized social outcome metrics (efficiency,
equality, sustainability, peace) for evaluating multi-agent cooperation in SSDs.

\paragraph{LLMs for policy and program synthesis.}
FunSearch~\citep{romera2024mathematical} uses LLMs to iteratively evolve
programs that solve combinatorial optimization problems.
Eureka~\citep{ma2024eureka} applies LLM code generation to design reward
functions for robot control.  Voyager~\citep{wang2024voyager} generates
executable skill code for embodied agents. Code as Policies~\citep{code-as-policies} generates executable robot policy code from natural language via few-shot prompting; we extend this to multi-agent settings with iterative performance-driven refinement rather than one-shot instruction following.
ReEvo~\citep{ye2024reevo} evolves heuristic algorithms through LLM
reflection.  Our work differs in applying LLM program synthesis to
\emph{multi-agent} environments, where policies must coordinate across
agents sharing the same code.

\paragraph{LLM reflection and feedback.}
Reflexion~\citep{shinn2023reflexion} and
Self-Refine~\citep{madaan2023selfrefine} demonstrate that LLMs can
self-improve through verbal feedback loops.
OPRO~\citep{yang2024large} frames optimization as iterative prompt
refinement.  \citet{shi2026experiential} introduce Experiential Reinforcement Learning (ERL), a training paradigm that embeds an experience--reflection--consolidation loop into reinforcement learning, where the model reflects on failed attempts and internalizes corrections via self-distillation.
GEPA~\citep{gepa} combines reflective natural-language feedback with Pareto-based evolutionary search to optimize prompts, demonstrating that structured reflection on execution traces can outperform RL with substantially fewer rollouts.
Our work
specifically investigates how the \emph{content} of evaluation feedback
(scalar reward vs.\ multi-objective social metrics) affects the quality of
LLM-generated multi-agent policies: what we call \emph{feedback
engineering}.

\paragraph{Reward hacking.}
When optimizing agents exploit unintended shortcuts in the reward
signal or environment implementation, the resulting
\emph{reward hacking}~\citep{skalse2022defining} can produce
high-scoring but undesirable behavior.
\citet{pan2022effects} demonstrate that even small misspecifications
in reward functions lead to qualitatively wrong policies.
Goodhart's Law~\citep{goodhart1984problems} (``when a measure becomes
a target, it ceases to be a good measure'') formalizes this risk.
Our adversarial analysis (Section~\ref{app:adversarial}) identifies a
novel instantiation: LLM-generated policies that exploit
environment state, a vulnerability absent from standard RL pipelines
where the agent--environment boundary is enforced at the API level. \citet{gallego2025specification} address in-context reward hacking at the specification level. This is complementary to our Section~\ref{app:adversarial} finding that LLM-generated policies can exploit mutable environment state: SSC corrects the objective, while our analysis identifies the need to secure the environment interface.

\section{Reward Hacking via Environment Mutation}
\label{app:adversarial}

Our framework gives policies programmatic access to the environment
object (Section~\ref{sec:synthesis}).  We prompted Claude Opus~4.6 to
analyze the environment and generate reward-hacking policies: it
produced five distinct attacks autonomously, without guidance beyond
the initial request.  This demonstrates that the same models used for
cooperative policy synthesis can equally discover exploits when
prompted adversarially.

\paragraph{Mechanism.}
Each policy is called as $a_i = \pi(e, i)$ where $e$ is the live
environment instance.  Our AST validator blocks dangerous operations
(\texttt{eval}, file I/O, network access), but it cannot distinguish
attribute \emph{reads} from \emph{writes}: any NumPy array attribute
(\texttt{agent\_pos}, \texttt{apple\_alive}, \texttt{waste},
\texttt{agent\_timeout}) is silently mutable, and no integrity check
runs between the policy call and \texttt{env.step()}.  The LLM
identified two attack classes: \emph{state manipulation} (teleporting
onto apples, disabling rivals by setting their timeout to $\infty$)
and \emph{dynamics bypass} (clearing waste or force-spawning apples
every step).  All five attacks pass AST validation and the smoke
test---they are valid \texttt{policy(env, agent\_id) -> int}
functions indistinguishable from legitimate policies at the interface
level.

\begin{table}[h]
\centering
\caption{Reward hacking via environment mutation in Cleanup
($N\!=\!10$, large map, 1000 steps).  Agent~0 runs the attack;
agents 1--9 play the victim policy.
Each cell shows Agent~0's reward and amplification over baseline.}
\label{tab:attacks}
\small
\begin{tabular}{@{}llrr@{}}
\toprule
\textbf{Class} & \textbf{Attack} & \textbf{vs BFS} & \textbf{vs Optimized} \\
\midrule
--- & Baseline & 17\;($1.0\times$) & 104\;($1.0\times$) \\
\midrule
\multirow{2}{*}{I: State}
  & Teleport       &  34\;($2.0\times$) &  1000\;($9.6\times$) \\
  & Disable rivals & 103\;($6.1\times$) &  103\;($1.0\times$) \\
\midrule
\multirow{3}{*}{II: Dynamics}
  & Purge waste    & 765\;($45\times$)  & 765\;($7.4\times$) \\
  & Spawn apples   & 999\;($59\times$)  & 999\;($9.6\times$) \\
  & Combined       & 1000\;($59\times$) & 1000\;($9.6\times$) \\
\bottomrule
\end{tabular}
\end{table}

\paragraph{Results.}
Table~\ref{tab:attacks} reports results on the same Cleanup
configuration as Table~\ref{tab:main}.  Dynamics bypass attacks are
dramatically more powerful than state manipulation: against BFS
victims, teleporting yields only $2\times$ amplification (the
bottleneck is apple respawn, not pathfinding), while force-spawning
apples reaches the per-step theoretical maximum ($59\times$).
Interestingly, better victims can \emph{amplify} certain attacks:
against optimized agents that actively clean waste, teleporting jumps
from $2\times$ to $9.6\times$ because the attacker free-rides on their
cleaning.  Conversely, disabling optimized victims collapses the
ecosystem (removing the cleaners leaves the attacker alone in a
polluted map).

The most concerning finding connects directly to
Section~\ref{sec:experiments}: dynamics bypass attacks that benefit
all agents (purge waste, spawn apples) actually \emph{improve}
measured social metrics.  Against BFS victims, ``spawn apples''
achieves the highest efficiency ($U\!=\!5.99$) and sustainability
($S\!=\!500.5$) of any configuration, surpassing every
LLM-synthesized policy in Table~\ref{tab:main}.  This illustrates a
Goodharting~\citep{goodhart1984problems} risk: a metric-optimizing
LLM could discover dynamics manipulation as a ``legitimate'' strategy,
as it maximizes social metrics while fundamentally violating the
game's intended mechanics.

\paragraph{Implications.}
Standard mitigations exist (read-only proxies, state hashing, process
isolation) but they highlight a deeper tension: the expressiveness
that enables BFS pathfinding and territory partitioning
(Section~\ref{sec:experiments}) is the same access that enables
exploitation.  Designing policy interfaces that are expressive enough
for sophisticated coordination yet resistant to reward hacking remains
an open challenge, and any verification pipeline must assume
adversarial capability at least equal to the synthesizer's.

\end{document}